\documentclass[letterpaper, 10 pt, conference]{ieeeconf}  

\IEEEoverridecommandlockouts                              
\usepackage{graphics} 
\usepackage{graphicx}
\usepackage[bottom]{footmisc}
\usepackage{nameref}
\graphicspath{{images/}}

\title{\LARGE \bf
Diversity of Ensembles for Data Stream Classification
}

\author{Mohamed Souhayel Abassi \\ abassi01@gw.uni-passau.de \\ Computer Science, University of Passau }

\begin{document}

\maketitle
\thispagestyle{empty}
\pagestyle{empty}

\begin{abstract}

When constructing a classifier ensemble, diversity among the base classifiers is one of the important characteristics. Several studies have been made in the context of standard static data, in particular, when analyzing the relationship between a high ensemble predictive performance and the diversity of its components. Besides, ensembles of learning machines have been performed to learn in the presence of concept drift and adapt to it. However, diversity measures have not received much research interest for evolving data streams \cite{DBLP:journals/inffus/KrawczykMGSW17}. Only a few researchers directly consider promoting diversity while constructing an ensemble or rebuilding them in the moment of detecting drifts. In this paper, we present a theoretical analysis of different diversity measures and relate them to the success of ensemble learning algorithms for streaming data. The analysis provides a deeper understanding of the concept of diversity and its impact on online ensemble Learning in the presence of concept drift.  
More precisely, we are interested in answering the following research question; Which commonly used diversity measures are used in the context of static-data ensembles and how far are they applicable in the context of streaming data ensembles?  

\vspace{\baselineskip}

\end{abstract}

\section{INTRODUCTION}

The collection, storage, and processing of data is considered to be one of the promising challenges in the field of data science. The massive volumes of data should be first collected and stored in different data sources, i.e characterized by the different speed at which data is passed to analytical systems.
 
In many environments where static data is used, analytics are performed on a data set that is not changing. However, data can also be continuously generated in the form of data streams which implies a new area of requirements and constraints such as processing time and memory capabilities.
Data streams pose new challenges for data mining and machine learning algorithms, such as typical batch learning for supervised classification which is not capable of efficiently analyzing the changes in the distribution of data which occur in the stream over time, called concept drift.  
This latter has been introduced to identify how data and target concepts change over time. It can lead in many cases to the deterioration of the predictive performance of the classifiers, especially when data comes so quickly that labeling all items may be delayed or sometimes even impossible.
 
Several algorithms were proposed, these are basically based on using sampling techniques, drift detectors, and sliding windows. One of the most promising research directions is ensemble learning, also known as the committee of learners, where each classifier in the ensemble is referred to as base classifier.
 
Recent decades have shown that ensemble methods are genuinely capable of incorporating new data by either introducing a new component to the base learners or update existing components. Thus, ensemble methods for block-based (batch-incremental) or online (instance-incremental) approaches, are quite naturally adapted to data streams.

Most of the previous studies \cite{DBLP:journals/tnn/BrzezinskiS14} on stream classifiers focus on the performance and computational costs of ensembles in different cases where concept drift was detected. Besides, the diversity of ensembles for static data has been recognized as a very important characteristic in classifier combination and many authors \cite{DBLP:journals/ijns/CarneyC00} \cite{DBLP:conf/nips/KroghV94} believe that the success of ensemble algorithms depends on both the accuracy and the diversity among the base learners \cite{DBLP:journals/aim/Dietterich97} \cite{DBLP:journals/ml/KunchevaW03}

However, despite the popularity of the term diversity, such interest is not visible in research of the role of diversity in the presence of concept drift, in scenarios where ensembles are fed by data streams. The primary goal of the study presented in this paper is to get a deeper understanding of how the diversity of ensemble classifiers can deal with their predictive accuracy in streaming scenarios. 

This paper is composed five parts, the first part details the data stream characteristics and introduces the data expiration problem caused by the evolving data. Part 2 lists seven different diversity measures applicable in the context of static-data ensembles and how they contribute in the improvement of the accuracy of the ensemble. Then we propose in part 3 an ensembling approach that is proved to be convenient to face the change of distribution in the streaming data. After that, an adaptive study of the term diversity is discussed in part 4 to conclude about how far are the diversity measurement applicable in the context of ensembles with evolving streams. Last but not least, part 5 propose a brief discussion of the relation of organic computing and the diversity of ensembles.

\section{DATA STREAMS AND CONCEPT DRIFT}
During the last few years, more sources of data have been installed and managed to get real-time data production, e.g GPS data, sensor networks, power grids, etc..  These diverse data sources produces a huge amount of data which should be labeled either in a fixed interval or in real time. 
In this chapter, a further understanding of data streams and concept drift is presented.

\subsection{Data Streams}

Data streams are basically an unbounded ordered sequence of data items whom events occur independently from each other.  The time arrivals between the arrival of each data may vary, a brief comparison between data streams and conventional static data sets is described in table \ref{table:1}

\begin{table}
\caption{Comparison between static data  and data streams \cite{DBLP:journals/inffus/KrawczykMGSW17}}
\label{table:1}
\begin{tabular}{ |p{2cm}||p{2.5cm}|p{2.5cm}|  }
 \hline
& Static Data	& Streaming Data\\
 \hline
 Data arrival	& No arrival	&Arrival on-line\\
 \hline
 Order control&	Order fixed & No control over order of arrival neither within nor across data\\
 \hline
 Size	&Bounded	& Unbounded \\
 \hline
 Queries	&One-time	&Continuous \\
  \hline
 \end{tabular}
 \end{table}

\begin{figure*}
    \center
    \includegraphics{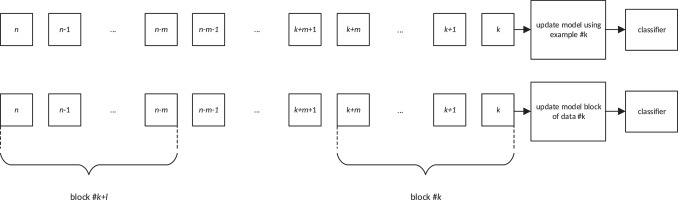}
    \caption{Table 1: Difference between incremental and block base classifier updating}
    \label{fig:fig1}
\end{figure*}

The table \ref{table:1} shows critical characteristics of the data streams where classic learning algorithms cannot be applied. For instance, batch learning which is not capable of fulfilling data stream memory and processing requirements. Also, incremental algorithms like Naïve Bayes and Neural networks are insufficient to tackle the nature of the data sources. Therefore, new techniques took birth such that windowing and sampling, but even these methods require maintaining all the elements in memory. 
An illustrative problem \cite{DBLP:books/daglib/0030859} of the streaming process is the problem of finding the maximum or minimum value is a sliding window over a sequence of numbers. In this case, whatever the window size, the first element in the window is always the maximum value, as the sliding window moves, we ought to maintain all the elements in memory to get the exact answer.
Examples from the data stream are provided either online or as data chunks, as depicted in figure \ref{fig:fig1}, these two modes have a crucial influence on the evaluation of classifiers.
\vspace{\baselineskip}

\subsubsection{Online data}
When the data is transmitted instance by instance, algorithms process single examples appearing one by one in consecutive moments in time. Hence, an arbitrarily online ensemble learning technique is applied in order to process each training example once on arrival without the need for storage or reprocessing and maintain a current hypothesis that reflects all the training examples so far \cite{DBLP:conf/kdd/OzaR01}.

\subsubsection{Data chunks}
\textit{Data chunks} also called data portions or data blocks, are large sets of data. Usually, they come with a fixed, equal size of a block in a way that the training and evaluation of the classifiers are done when all examples from a new block are available. In few cases, accumulating data for a certain period and labeling it is more reliable and genuinely applicable as when it is in an online setup.

\subsection{Concept drift}

Many studies \cite{Krawczyk:2017:ELD:3075743.3075861} agreed that there are basically two forms of data streams, the stationary; where examples are drawn from a fixed probability distribution and non-stationary which describes data that can evolve over time. In many cases, the target concepts, i.e. classes or labels, can also be affected and change in distribution.

The term \textit{concept} refers to the whole distribution of the data in a certain point in time as Narasimhamurthy and Kuncheva \cite{DBLP:conf/aia/NarasimhamurthyK07} has already defined. Being characterized by the joint distribution P(x,y) where x represents the sample input and y its label. Hence, the concept from which the data stream is generated shifts after a certain period of time which gives birth to the phenomena concept drift, also called \textit{covariant shift}. In other words, for each point in time t; $ P_{t}\big(x,y\big)  \neq P_{t+ \Delta }\big(x,y\big) $

Different components of the joint probability may change, this can be translated by the change of feature selection. In particular,  when concept drift occurs either on one or both of the of the prior probabilities of classes P(x) or the class conditional probability P(y/x). Thus, we distinguish two types of drifts: real drift and virtual drift \cite{DBLP:conf/ecml/WidmerK93}

\vspace{\baselineskip}
\subsubsection{Real drift}
\textit{Real concept drift} refers to changes in P(y/x), that means a change in the class boundary. Such changes can occur with or without changes in P(x) and can be visible in cases where knowing true class labels is not evident.

\subsubsection{Virtual drift}
In \textit{virtual concept drift}, the distribution of instances may change, which corresponds to changes in the class priors of the classes but the underlying concept does not vary. In such cases, the probability distribution may change the error of the learned error, this can only be relevant when the data stream exhibits class imbalance.

\vspace{\baselineskip}
Current researches related to learning classifiers for stream data focus mostly on real drifts since few techniques that were introduced to handle real drifts may still work for certain sub-categories of virtual drifts. Apart from the causes and effect of concept changes, researches \cite{DBLP:journals/inffus/KrawczykMGSW17} propose further characteristics such that their performance, frequency, and the predictability
Many studies and application-oriented papers, such as \cite{Žliobaitė2016},proved that the problem of concept drift must be recognized and addressed in multiple application areas, This is an evidence of the strong requirement for streaming classifiers, respectively streaming ensembles, to be able to predict, detect and adapt to concept drifts.

\subsection{The data expiration problem}
Learning drifting concepts is the origin of a fundamental problem called data expiration problem, this latter occurs during mining of concept-drifting data streams. In other words, learning data that change in distribution over time makes identifying them in the training set no longer an easy task, due to the fact that the data is no longer consistent with current concepts.
A simple proposed solution is done by discarding the data indiscriminately after a fixed period of time, noted T, after their arrival, the data is therefore considered old. On one hand, if the period T is large, a vulnerability to unpredict conceptual changes in the data takes place and the classification accuracy is respectively reduced. 
On the other hand, if T is small, The data may be not enough for the training set and results in a higher variance since the learning model is over-fitting.

\begin{figure}
    \centering
    \includegraphics[width=\linewidth]{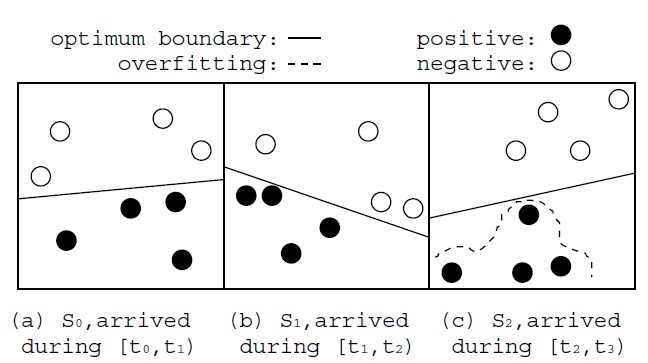}
    \caption{Optimum boundary for sequential arrival \cite{DBLP:conf/kdd/WangFYH03}}
    \label{fig:fig2}
\end{figure}
\begin{figure}
    \center
    \includegraphics[width=\linewidth]{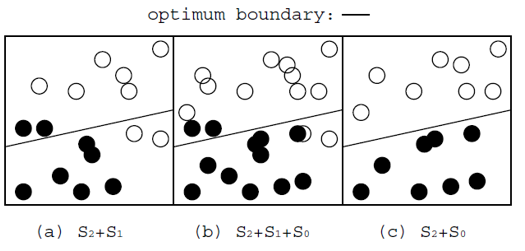}
    \caption{Optimum boundary for random arrival \cite{DBLP:conf/kdd/WangFYH03}}
    \label{fig:fig3}
\end{figure}

To simplify the problem, a 2-dimensional data is partitioned into sequential chunks based on the moment of arrival. The figures \ref{fig:fig2} and \ref{fig:fig3}{} below shows the distribution of the data  S =\{$ S_{1} $,...,$ S_{n} $\} that arrives between  $ t_{i} $ and  $ t_{i+1} $ respectively and the corresponding optimum decision boundary during each time interval.
For instance, if the data $ S_{3} $ arrives at a time $ t_{4} $, which part of the training set should be retained to make the feature plan, in the current model so that data that is arriving at t4 can be classified as correct as possible?

Figure \ref{fig:fig2} shows that using the most recent data in the stream as the training set is one of the ways to handle new appearing concept, but it will not avoid over-fitting; Here we observe that the learned model is over-fitting.
Figure \ref{fig:fig3} shows that using more historical data in training will eventually solve this problem but reduces accuracy.

Recent research study \cite{DBLP:conf/kdd/WangFYH03} showed that data should not be discarded based on time of arrival, rather the class distribution. In other words, historical data where classes are distributed similarly to the current streaming data are the crucial elements that can reduce the variance of the current model, avoid over-fitting from occurring and increase predictive accuracy.

In a realistic scenario, an ensemble of classifiers is proved
to be helpful to solve the data expiration problem 
This means, that compared with a single classifier, which gets the training data in windows form of the k blocks, the classifier ensemble approach is capable of reducing classification error in scenarios where classifier's weight is considered to be reversely proportional to its expected error. Such a weighting scheme should be assigning classifiers representing totally conflicting concepts near-zero weights.
The next section lists a number of diversity measures that are elligible to show how 'diverse' an ensemble of learners is.

\section{DIVERSITY IN STATIC-DATA ENSEMBLES}

Since there exists no unique definition of the term diversity of an ensemble, we will illustrate few basic measures of diversity that can be used as a promising hint about how diverse ensemble classifiers are, for scenarios where data is static and its distribution does not change in time. Therefore, the diversity is basically modeled in a bias of the statistical and pairwise/non-pairwise diversity measures.
The following measures were proposed by different studies independently.  
 
To illustrate the calculation of each measure, few parameters should be first defined for the ease of comprehension. We consider a data-set of N data samples, noted $ x_{i} $ , each have a class label $  y_{i} $. The L base classifiers defined by the set $ H =\{h_{1},h_{2},...,h_{L}\} $ are trained with the training set $ Tr=\{(x_{1},y_{1}),(x_{2},y_{2}),...(x_{N},y_{N})\}. $
Each classifier j has an output for a  sample xi is $ h_{j}(x_{i}) $ , which corresponds to the weight $ w_{j} $ in the ensemble set of weights with a classification accuracy equal to $ p_{j} $.
The correct/incorrect decision is a N * L matrix, also called Oracle output, noted O, whose elements are defined in $ \{-1,1\} $. In other words, $ o_{ij} = 1 $ if training sample $ x_{i} $ is classified correctly by the base classifier $ h_{j} $, -1 otherwise. 
Clearly, Oracle outputs are only possible for a labeled dataset and they provide a general model for analyzing a classifier ensemble, this model can genuinely draw generalized conclusions applicable within various ensemble learning methods.
 
The average classiﬁcation accuracy of the base classifiers on the training data, noted P, is defined as follows :
$$ P = \sum_{j = 1}^L  w_{j}  p_{j}  \eqno{(1)}
$$
$$
P = 1 -  \frac{ \sum_{i = 1}^N l_{i} }{NL} \eqno{(2)}
$$

Where $ l_{i} $ illustrates product of L and sum of the weights of the base classiﬁers that classify the training sample $ x_{i} $ incorrectly, defined as :
$ l_{i} = L \sum_{O_{ij}=-1}  w_{j} $

As we mentioned, the ensemble is composed of L different base learners. This latter produces $ \frac{L(L-1)}{2} $ pairwise diversity values. For simplicity purposes, the study in this paper focuses on ensembles that are composed of two component classifiers $ C_{1} $ and $ C_{2} $. Also, the weighted average voting method is used for ensembling the predictions, where the ensemble classifies a new sample with the class label that has the highest weighted vote of the base classifiers.
 
Let us consider these two aforementioned base classifiers’ outputs and resume the results in a 2 * 2 matrix as shown in table \ref{table:2}:
 
\begin{table}
\centering
\label{table:2}
\caption{The 2x2 ensemble component relationship table with probabilities}
 \begin{tabular}{||c c c||} 
 \hline
 & $ C_{i} $ correct  & $ C_{i} $ wrong\\ [0.5ex] 
 \hline\hline
 $ C_{j} $ correct & a & b \\ 
 \hline
 $ C_{j} $ wrong & c & d\\
 \hline
 
 \end{tabular}
\end{table}

The table \ref{table:2} presents proportions of correct/incorrect answers of one of or both components, thus, the total of all the cell values a+b+c+d = 1.

A proposed approach to distinguish between the different diversity is to take into consideration the elements that contribute for every measure. In other words, there is basically two categories of diversity measures; pairwise and non-pairwise measures .

\subsection{Pairwise diversity measures} 
Four of the well-know pairwise diversity measures are discussed. That means, every measure can be applied on a pair of base learners which will produce $ \frac{L(L-1)}{2} $ pairwise diversity values. To get a single one value it is necessary to average across all parts that compose the ensemble

\subsubsection{The Correlation Coefficient $ \rho $ }
One of the trivial measures that can be calculated directly from the aforementioned table is the correlation between two binary classifier outputs, for instance, y1 and y2.
 
Correlation can be calculated for a pair of Oracle outputs, their values are illustrated in table \ref{table:2} as the probabilities for the respective pair of correct/incorrect outputs. Hence, who is calculated as follows :
$$
\rho_{1,2} =  \frac{ad-bc}{\sqrt{(a+b)(d+c)(b+d)(a+c)}} \eqno{(3)}
$$
Breiman \cite{DBLP:journals/ml/Breiman01} derives an upper bound on the generalization error of random forests using the averaged pairwise correlation, which also demonstrates that lower correlation leads to better ensembles.

\subsubsection{Q statistics}
There are various statistics to assess the similarity of two classifier outputs \cite{DBLP:books/daglib/0081942}. Using table \ref{table:2}, Yule’s Q statistics for two classifiers, $ C_{1} $ and $ C_{2} $ is :
 $$ Q_{1,2} =  \frac{ad-bc}{ad+bc}  \eqno{(4)} $$
The equation (5) reflects intuitively that Q and $ \rho $ have the same sign and $ |\rho|   \leq  |Q| $.
In scenarios where the classifiers are statistically independent, the correspondent prior probabilities are equal to actual probability, this leads to a value of $ Q_{1,2} = 0 $. In other cases, Q varies between -1 and 1, e.g negative if classifiers commit errors on different objects and positive if they tend to identify the same class label  correctly.
 
In general, we tend to ensemble L classifiers to average the weighted vote and consider a set of pairs of classifiers to facilitate the calculation of Q statistics. Hence, the average Q statistics over all pairs of classifiers is:

$$ Q_{av} =  \frac{2}{L(L-1)}   \sum_{i=1}^{L-1}  \sum_{j=i+1}^L Q_{i,j} \eqno{(5)} $$

\subsubsection{The Disagreement Measure}
The disagreement measure is basically defined as the probability that two diverse classifiers perform differently on the same training data . It  was first proposed by Skalak \cite{DBLP:conf/dmkd/Skalak01} in order to evaluate the diversity between two base classifiers and then employed by Ho \cite{Ho:1998:RSM:284980.284986} as a diversity measure in a decision forest for bagging method.
In other words, It is the ratio between the number of observations on which one classifier is correct and the other is incorrect to the total number of observations. Hence, the diversity increases with the value of the disagreement measure.
According to the table \ref{table:2}:
$$ dis_{1,2} = b + c \eqno{(6)} $$

For a set of L classifiers, this diversity measure is calculated as the average value over all pairs of base classifiers, in our notation :

$$ dis_{av} =  \frac{2}{L(L-1)}   \sum_{i=1}^{L-1}  \sum_{j=i+1}^L dis_{i,j} \eqno{(7)} $$

\subsubsection{The Double-Fault Measure}
Another intuitive choice to measure the diversity within an ensemble is the double-fault value which is based on the fact that it is more important to detect when the simultaneous errors are being committed than when both classifiers are correct. In other words, illustrates the probability that the two classifiers $ C_{1} $ and $ C_{2} $ both act wrong during the prediction process, i.e the proportion of the cases that have been misclassified by both classifiers In out notation 

$$ DF_{1,2} = d \eqno{(8)} $$

This measure was proposed by Giacinto and Roli \cite{Giancinto01designof} in order to select classifiers that are least related from a pool of classifiers. These two researchers claimed that the fewer occurring coincident errors between two classifiers, the more diverse they are.
Ruta and Gabrys \cite{Ruta01analysisof} showed that one of the properties of the DF measure is the non-symmetry, i.e if we swap the -1s and 1s, DF will no longer have the same value

Same as previous diversity measure, the DF measure for a set of L classifiers is calculated as follows :

$$ DF_{av} =  \frac{2}{L(L-1)}   \sum_{i=1}^{L-1}  \sum_{j=i+1}^L DF_{i,j} \eqno{(9)} $$

\subsection{Non-paiwise diversity measures}
This part of the paper is dedicated to introducing the most common measure where all classifiers are considered as a whole buck so that the result concerns directly only the ensemble and is related to the oracle classifiers’ outputs that are assumed again as aforementioned, i.e. 1 for correctly labeled samples and -1 otherwise.

\subsubsection{The Entropy Measure E}
The Entropy measure is based on the concept that the highest diversity among a set of L classifiers for a particular $ x_{j} $ is manifested by  [L/2] \footnote{the ceiling function and returns the smallest integer greater than a. } of the votes with the same value ; either -1 or 1, and the other L- [L/2] alternative value. For instance, if they all were -1s or all were 1s, there is no disagreement and the diversity is at its minimum value 0.

$$
E =  \frac{2}{L-1}  \sum_{i=1}^N min( \sum_{j=1}^L {h_{j}(x_{i})}, L - \sum_{j=1}^L h_{j}(x_{i} ) ) \eqno{(10)} $$

Kuncheva  \cite{DBLP:conf/kdd/WangFYH03}  mentioned that E is not a standard measure because it does not use the logarithm function. On the other hand, Cunningham and Carney \cite{DBLP:conf/ecml/CunninghamC00} proposed a more specific entropy measure, denoted Ecc, in which the to researchers assumed that the number of the classifiers tend to infinity ${L\to\infty}$, the entropy is then calculated, for a C proportion of correct outputs, as follows :
$$ Ecc = -a log(a) –(1-a) log(1-a) \eqno{(11)} $$
Add to that, these two aforementioned quasi-similar measures have a similar pattern of relationship with the ensemble accuracy.

\subsubsection{Kohavi-Wolpert Variance}
The concept of bias-variance trade-off has generated new decomposition formula of the classification error of a  classifier. This latter was proposed by Kohavi and Wolpert \cite{DBLP:conf/icml/KahaviW96} They give an original expression of the variability of the predicted class label y for x, across training sets, for a specific classifier model $ C_{j} $ :

$$
Variance_{x}=  \frac{1}{2} ( 1 -  \sum_{i=1}^c {{P(y=w_{i} /x)}^{2}} ) \eqno{(12)} $$

Instead of C classes, in this paper we consider C=2 possible classifier outputs: correct and incorrect

Kuncheva and Whitaker presented a modified version of the equation below;
$$
KW = \frac{1}{N{L^{2}}} \sum_{i=1}^N l_{i}(L-l_{i}) \eqno{(13)}
$$
It is important to note that the diversity increases with values increasing of the KW variance

\subsubsection{Inter-rater Agreement}
Fleiss \cite{Fleiss} has developed another diversity measure related to inter-rater reliability, denoted  $ \kappa $, which can be used to measure the level of agreement within a specific set of classifiers. Thus, the diversity increases when the classifiers disagree with one another, i.e. the value of  $ \kappa $ decreases. The inter-rater agreement is calculated as follows:
$$
\kappa = 1 -  \frac {\sum_{i=1}^N {{l_{i}(L-l_{i})} } } {NL(L-1)P(1-P)} \eqno{(14)} $$

\subsubsection{the Generalized Diversity Measure}
Partidge and Krzanowski \cite{PARTRIDGE1997707}proposed the Generalized diversity measure, denoted GD. They argued that maximum diversity is achieved when the failure of one classifier is
accompanied by correct classification by the other classifier and minimum diversity occurs
when two classifiers fail together. According to t the previous notation, for a random training sample $ x_{i} $ :
$$ 
GD = 1 -  \frac{\sum_1^L { \frac{j(j-1)}{L(L-1)} T_{j} }}{ \sum_1^L { \frac{j}{L}    T_{j} }                    }
\eqno{(15)} $$

Where $ T_{j} $ denote the probability that $ l_{i} $ = j, i.e. the probability that exactly j out of the L classifiers fail on a randomly chosen input.
The next section includes a forward discussion on how such ensembles are built and how their diversity can affect accuracy.

\section{STREAM DATA ENSEMBLES}
Usually, the primary task of a learner is to predict a discrete a class label for classification problems or numeric output values for regression problems. As most of the current research on data stream ensembles concerns classification, this section will be dedicated to the classification research direction. However, nearly all of the above-mentioned points are also valid for regression cases.

Stream data ensemble is basically a combination of committees that learn from stream data within one of the three major frameworks; supervised learning, semi-supervised learning or unsupervised learning.

The majority of proposed algorithms for learning stream classifiers follow the supervised framework, i.e. with frequent and complete access to class labels for all processed examples. This assumption is not realistic since the class labels in data streams are naturally not immediately available in scenarios where a class is only known after a long delay. For instance, fraud detection and credit approval applications.

Therefore, researches considered other frameworks as alternatives to the classic approaches where delay assumption is absent or quasi-negligible;

Hofer and Krempl \cite{DBLP:conf/icdm/KremplH11} proposed a learning approach with delayed labeling when access to true class labels is available after an unknown period of time. The classifiers may then be able to the stream earlier without knowing it.
Another approach proposed by  Dyer, capo, and Polikar \cite{6604410} consists of learning from initially labeled samples where an initial classifier learns from a limited number of labeled training samples and then processes the upcoming stream of unlabelled samples without any access to their labels.

From all, a popular and efficient way to adapt to data drift is using accuracy-weighted ensembles, through which we assign an actual weight to each classifier that reflects its predictive accuracy on the current testing data.

\subsection{Accuracy-weighted ensembles}
The base assumption, as Scholz and Klinkenberg \cite{DBLP:journals/ida/ScholzK07} explained, is to feed an ensemble of classifiers with a data generated from a mixture of distribution which can be considered as a weighted combination of distributions characterizing the target concepts, this process occurs during the change.

Wang, Fan, Yu, and Han \cite{DBLP:conf/kdd/WangFYH03} proved, through an error reduction concept, that compared with a single classifier that learns from the examples in the entire window of k chunks, the classifier ensemble would perform better by reducing classification error through the weighing scheme. They considered a classifier’s weight as a reverse proportional value of the expected error. However, it has been mentioned that the aforementioned property is accompanied with a weighting scheme that assigns classifiers near-zero weights if they represent totally conflicting concepts.

Street and Kim \cite{DBLP:conf/kdd/StreetK01} introduced one of the first approaches, called SEA Algorithm. SEA builds separate classifiers from sequential chunks of training examples, which will be combined a later step into a fixed-size ensemble using heuristic replacement strategy.

Another interesting algorithm was introduced by Wang, Fan, Yu, and Han, which uses the main approach in a way that combining the learners is based on a judiciously weighting using their expected classification accuracy on the test data under the time-evolving conditions. The classification models used for this purpose are basically C4.5, RIPPER, naive Bayes, and others. These models, applied in the context of this approach, has shown an improvement in two scales, the efficiency in learning the model and the accuracy in performing classification.

Let us assume that the data stream can be partitioned into sequential chunks, noted $ S_{1} $ the first chunk to $ S_{n} $ as the last and newest incoming chunk. To every chunk, a correspondent classifier $ C_{i} $ is assigned to learn from this chunk. It is important to note that we use a minimum of two classifiers to build the ensemble.

According to the error reduction property that was aforementioned, every classifier is expected to have a weight reversely proportional to the estimated error of its proper prediction on the test examples. The class distribution of $ S_{n} $ is then assumed to be the closest to the class distribution of the current test data. Therefore, the classifiers weights issue can be minimized to a computation of the classification error on $ S_{n} $. 

In other words, if $   f_{c}^{i} (x) $ is the the probability given by $ C_{i} $ that x is an instance of class c, then the mean square error of the latter classifier can be expressed as follows :
$$
 MSE_{i} = \sum_{(x,c) \epsilon  S_{n}}   \frac{{(1-f_{c}^{i} (x))^{2}} }{ \mid  S_{n} \mid }
 \eqno{(16)} $$

Analogically, if we consider a random classifier with a p(c) as the class distribution,i.e the probability of observing the class c, then its mean square error is expressed by:
$$
 MSE_{r} = \sum_{c}   {p(c)(1-p(c))^{2}}
 \eqno{(17)} $$

Thus, the weight $ w_{i} $ of the classifier is , as previously explained expressed by : $ w_{i} = MSE_{r} - MSE_{i} $

\begin{table}
\caption{benefit matrix $ b_{c,c’} $}
\label{table:3}
\centering
\begin{tabular}{ |p{2cm}||p{2.5cm}|p{2.5cm}|  }
 \hline
 & predict fraud	& predict $ \neg $ fraud\\
 \hline
 actual fraud	& t(x) - cost	& 0\\
 \hline
 actual $ \neg  $ fraud &	- cost & 0\\
  \hline
 \end{tabular}
 \end{table}

Few applications like credit card fraud detection are cost-sensitive. Wang, Fan, Yu, and Han proposed a shortcut approach by using the benefits achieved by a certain classifier on the most recent training data block as its weight. The benefits matrix illustrated in table \ref{table:3}, called $b_{c,c’}$ that shows the benefit of classifying transaction x of actual class c as an instance of another class c’, noted $ b_{c,c’}(x) $.

\begin{itemize}
\item t(x): the transaction amount
\item Cost : the fraud investigation cost
\end{itemize}
To sum up, the total benefits achieved by a classifier $ C_{i} $ is expressed by : 
$$
b_{i} = \sum_{(x,c) \epsilon  S_{n}}   \sum_{c'} {b_{c,c’}(x) . f_{c}^{i} (x) }
 \eqno{(18)} $$
 
In analogy to the general approach, if $ b_{i} $ is the benefits achieved by a classifier that predicts randomly, the weight of the classifier $ C_{i} $ is here expressed as follows: $$ w_{i}=b_{i}-b_{r} \eqno{(19)} $$
 
However, with a huge amount of data streams, there is a risk to have an infinite number of classifiers. Therefore, tend to use only the top N classifiers that manifest the highest prediction accuracy on the current training data stream. 

Wang, Fan, Yu, and Han identified an ensemble pruning technique to select the set of classifiers that can build the ensemble for a better predictive performance in scenarios where concept-drifting data streams compose the training data-set.

\subsection{Algorithm and Complexity}
\begin{figure}
    \centering
    \includegraphics[width=\linewidth]{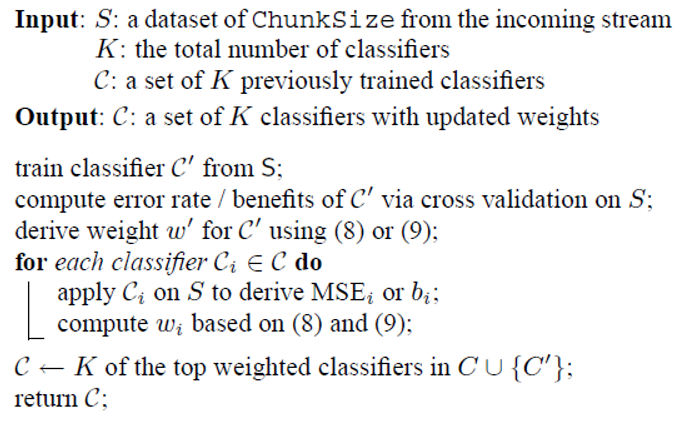}
    \caption{Algorithm 1: A classifier ensemble approach for mining concept-drifting data streams \cite{DBLP:conf/kdd/WangFYH03}}
    \label{fig:fig4}
\end{figure}
Theoretically, the algorithm 1 depicted in figure \ref{fig:fig4} summarizes the aforementioned approach related to an average-weighted ensemble for mining concept-drifting data streams and the selection property of the best performing classifiers.
This algorithm outlines that whenever a new chunk of data has arrived, a classifier is built based on this chunk and a new weight is attributed to it with respect to the weights tune of the previous classifiers.
Practically, the chunk size is quite small to allow an ease process of data stored in memory. 

The complexity of this algorithm in classifying test examples is linear in the size of the test data set and expressed, in the aforementioned notation, as follows: $ O(n . f(\frac{s}{n}) + K_{s}) $
where we note f(s) as the complexity for building a classifier on a data set of
size s, this function is usually super-linear, which signifies that the ensemble approach presented is more efficient.
In the next section, we will adapt these measures to the case of ensembles with evolving data streams.

\section{DIVERSITY IN DATA STREAM ENSEMBLES}
As previously mentioned, diversity is one of the crucial evaluation elements of ensemble-based classifiers. According to the error reduction property, an ensemble can perform better than a unique one, in case the prediction process is quite diverse.
Although few research studies, for instance, the study of Kuncheva and Whitaker \cite{DBLP:journals/ml/KunchevaW03} were conducted to show correlations between accuracy and specific diversity measures for some special cases
The main issue is actually to prove how the augmentation of a certain diversity measure can lead to an augmentation of the overall accuracy. It becomes even more complicated since there is no generally accepted definition of diversity.

For scenarios where training data is composed of data streams, diversity has been partially studied. A recent study was driven by Brzezinski and Stefanowski \cite{DBLP:conf/dis/BrzezinskiS16} presents ways of calculating diversity, visualizing them with histograms and introduced a technique to use them for drift detection as additional information from the mainstream.

\subsection{Diversity Techniques}
Many diversity definitions assume that learners of the ensemble cannot be random guessers nor learners with different internal representations that consistently predict the same class labels.
Gomes,Barddal, Enembreck and Bifet \cite{DBLP:journals/csur/GomesBEB17} suggested few elementary techniques that can be relevant to induce more diversity to an ensemble, either by manipulating the input data, the underlying classifiers, and heterogeneous base learners, and the output predictions;
\vspace{\baselineskip}
\subsubsection{Input Manipulation}
many common input manipulation methods have been granted to induce diversity, this includes dividing the stream data into chunks and feed them to the training classifiers It is also possible to train the classifiers on different subsets of features instead of different subsets of instances, this strategy is called Random Subspace Method (RSM). For a feature space of m dimensions, there are basically 2m – 1 different non-empty subsets of features, thus it is more probable to be infeasible to train one learner for each subset given a high dimensional dataset.
This is why RSM is usually associated with decision trees. In fact, training ensembles using RSM yield several advantages such as diversity enhancement and better predictive performance. Many pieces of research have been conducted in this area, and have introduced several RSM sub-techniques. For instance, Streaming Random Forests, dynamic streaming random forests, restricted Hodding Trees, etc..

\subsubsection{Base learner manipulation}
Another promising technique to achieve diversity is to modify the characteristic of every base model, For instance, implement multiple decision trees with different topologies or with the same topology but starting with different weights at the first layer. Many authors like Bifet and Gavalda \cite{Bifet:2009:ALE:1617420.1617445} propose the Adaptive Size Hoeffding Trees (ASHT) Which based on the bagging algorithm. This latter is an ensemble of decision trees of varying sizes. The authors proved that smaller trees are able to adapt rapidly to drifts, while bigger trees are useful during stable periods.

\subsubsection{Heterogenous Base Learners}
it is very convenient to parameters of the same base learners and obtain ensemble classifiers with different biases. Examples of algorithms implementing this approach are HEFT-Stream \cite{DBLP:conf/pakdd/NguyenWNW12} and HSMiner \cite{DBLP:conf/ictai/ParkerMK12}
For instance, HEFT-Stream trains heterogeneous learners on different samples and subspace of data, it keeps the same ensemble learners and if a sudden drift takes place it adds a new learner whose base learner matches the current learner with the highest weight.

\subsubsection{Output manipulation}
Last but not least, the manipulation of the output method which is basically presented for multi-class classification rather for binary since the latter is quite easy to manipulate. It has been known that a One-Versus-All  approach is a promising approach since it decomposes the original problem into multiple binary problems and assigns different classifiers to each class. This latter proves how diversity can be increased proportionally to a by-product.
However, this approach should be combined with Error-Correcting Output Codes (ECOC) in order o encounter the problem of the high update cost and imbalanced class distributions.
Examples of such algorithms include One-Versus-All Decision Trees, HSMiner.
To assess how effective one diversity inducing technique is, one could choose to observe the ensemble overall predictive accuracy. In the next section, we present a few diversity measuring techniques, but this time in a data stream setting.

\subsection{Diversity Measures for Streaming Data}
Although high diversity may not directly indicate high accuracy, measuring diversity can be useful to analyze the effectiveness of the aforementioned diversity inducing methods. 
For the static-data distribution scenario, we distinguished between the pairwise and non-pairwise diversity measures. Here, another categorization is going to be introduced.
A recent study conducted by Brzezinski and Stefanowski \cite{DBLP:conf/dis/BrzezinskiS16} showed a possibility to calculate ensemble diversity measures in three basic stream processing scenarios; in chunks, incrementally and prequentially.

\subsubsection{Block-based processing}
Also called chunk-based processing, is the most trivial case where examples arrive in blocks of fixed size. The size is known for the evaluator, this means that the ensemble diversity measuring process on each incoming data block is quite a similar tot he one conducted for static data. In other words, the predictions are distinct on each block and the static-data diversity measure would be then applied in the form of pair-wise component relationships previously mentioned. Thus, for a data block of some examples and a set of ensemble classifiers, the six measures previously listed can be computed and costs constant time and memory per block.

\subsubsection{Incremental Processing}
it is slightly less trivial as the previous strategy, based on the assumption that a measure can be computed based on a summary of all previous examples and a single new example. But, it is possible to introduce an updated table of the a,b,c and d counts that illustrate the new incoming instance effect. 

\subsubsection{Prequential Processing}
It consists of incrementing and forgetting, this means, if the stream is subject to changes, the diversity can be then measured using two basic approaches to calculate values and store them temporarily; sliding window and the fading factors. In one hand, Sliding windows can limit the number of analyzed examples by retraining only an arbitrarily set of the recent records at each point of time. 
Thus, a single sliding window can be seen as a set of data blocks whose diversity measures are the same as the first listed processing. It is important to note that updating is important in calculating the diversity measures since sliding windows work with forgetting and storing new example after a certain period of time. 
On the other hand, Fading factors is performed on a stream of objects x by discarding the old records across time t through a multiplication of the incoming example, correspondent to a fading sum $ S_{x,\alpha} $ by a factor $ \alpha $. Then, a new value is added and computed as the incoming example, correspondent to a fading increment $ N_{\alpha} (t)  $
$$
S_{x,\alpha}(t) =  x^{t} + \alpha * S_{x,\alpha}(t-1)
\eqno{(20)} $$
$$
N_{\alpha} (t) = 1 + \alpha * N_{\alpha}(t-1)
\eqno{(21)} $$
In analogy to the scenario of static data, diversity measures are calculated when take in account that x can be counts of any of the values a,b,c or d in table \ref{table:2}.
For example, if both components misclassify and example at a time t, then double fault measure can be calculated as 

$$
DF_{\alpha}(t) = \frac{S_{d,\alpha}(t)}{N_{\alpha} (t)}
\eqno{(22)} $$

To sum up, if it is either static or stream data, diversity measures can be expressed and calculated with a minimal adaptation to the processing approach, while fulfilling limited time and memory requirements of stream processing. Prequential and incremental processing, for instance, require $ O(L^{2}) $ time and $ O(L^{2}) $ memory. 

\section{ENSEMBLE LEARNING IN ORGANIC COMPUTING SYSTEMS}
Ensemble learning can be related to organic computing \cite{DBLP:journals/corr/abs-1808-05443} through the fact this last form a sort of ensemble of machines that can cooperate with each other and evolve over time. In such systems, each individual – equivalent to a learner in ensemble learning – is probably autonomous and viewed as a whole within the committee, can respond dynamically to changes in the environment and can also have the sufficient freedom to do so.
Such organic computing-based systems can be set to a data that evolves over time and give birth to a change in the distribution of the training examples. 

Synonymous as when dealing with data stream mining, traditional machine learning methods are not very robust. Hence, using ensemble learning algorithms which are capable to learn from streaming
data, can help OC systems to tackle the above-mentioned challenges and at the same time improve their performance. 
Thus, the diversity of such ensembles plays an important role in having a new overview of the performance of the ensemble and the way it can be improved.

Authors in \cite{DBLP:journals/corr/TomfordeSM17} explain the main properties of organic systems. Such systems can be modified at runtime in terms of structure and/or behavior, this is quite the case of ensembles in data streams when we tend to make them be ‘more diverse’. In other words, the base learner architecture can be modified as discussed in part 4. Such an operation can also be described as a mechanism of configuration and management in order to change the behavior of the ensemble and improve predictive performance.

\section{CONCLUSION AND FUTURE WORK}
When it comes to data stream learning, adaptive models are then used to target training examples where class distribution changes over time. However, these models need to be maintained up-to-date to high amounts of data arriving at high speeds using limited resources. Ensemble-based methods have shown to be suitable to cope with data streams as they achieve high accuracy and can be combined with respect to the average-weighted approach in order to address issues such as concept drifts. This paper presents the main characteristics of data streams, ensemble learners, identifies the diversity measure for static-data ensembles; and discusses their equivalent measure for stream-data ensembles. 
In this work, we consider a weighted ensemble classifier on concept-drifting data streams. It combines multiple classifiers weighted by their expected prediction accuracy on the current test data. 
A diversity analysis with and without concept drift problems is also presented. The analysis shows that diversity of ensembles of evolving data can be measured as assumptions on which strategy is being employed during the training, among incremental, prequential or block-based processing.
Future works include further study of diversity in different types of concept drifts such as recurrent drifts, an adaptation of the ensemble classification setting to the challenging real-world scenarios, especially semi-supervised and imbalanced data streams, in which the flexibility of ensembles is quite appreciated.

\bibliographystyle{plain}

\bibliography{refs}

\end{document}